\documentclass[letterpaper]{article} % DO NOT CHANGE THIS

\usepackage{aaai2026}  % DO NOT CHANGE THIS
\usepackage{times}  % DO NOT CHANGE THIS
\usepackage{helvet}  % DO NOT CHANGE THIS
\usepackage{courier}  % DO NOT CHANGE THIS
\usepackage[hyphens]{url}  % DO NOT CHANGE THIS
\usepackage{graphicx} % DO NOT CHANGE THIS
\usepackage{booktabs}
\usepackage{multirow}
\usepackage{amsfonts}
\usepackage{natbib}  % DO NOT CHANGE THIS AND DO NOT ADD ANY OPTIONS TO IT
\usepackage{caption} % DO NOT CHANGE THIS AND DO NOT ADD ANY OPTIONS TO IT
\usepackage{algorithm}
\usepackage{algorithmic}
\usepackage{xcolor}

\usepackage{amsmath}
\usepackage{booktabs}
\usepackage{multirow}
\usepackage{newfloat}
\usepackage{listings}
\DeclareCaptionStyle{ruled}{labelfont=normalfont,labelsep=colon,strut=off} % DO NOT CHANGE THIS
\floatstyle{ruled}
\newfloat{listing}{tb}{lst}{}
\floatname{listing}{Listing}

\title{GraphFedMIG: Tackling Class Imbalance in Federated Graph Learning via Mutual Information-Guided Generation}

\author {
    Xinrui Li\textsuperscript{\rm 1},
    Qilin Fan\textsuperscript{\rm 1*},
    Tianfu Wang\textsuperscript{\rm 2},
    Kaiwen Wei\textsuperscript{\rm 3},
    Ke Yu\textsuperscript{\rm 1},
    Xu Zhang\textsuperscript{\rm 4}
}
\affiliations {
    \textsuperscript{\rm 1}School of Big Data and Software Engineering, Chongqing University\\
    \textsuperscript{\rm 2}AI Thrust, The Hong Kong University of Science and Technology (Guangzhou)\\
    \textsuperscript{\rm 3}College of Computer Science, Chongqing University\\
    \textsuperscript{\rm 4}School of Electronic Science and Engineering, Nanjing University\\
    20214093@stu.cqu.edu.cn,
   fanqilin@cqu.edu.cn,
   tianfuwang.cs@gmail.com,
   weikaiwen@cqu.edu.cn,\\
   keyu@cqu.edu.cn,
   zhang17@nju.edu.cn
}

\usepackage{bibentry}
\begin{document}

\maketitle

\begin{abstract}
Federated graph learning (FGL) enables multiple clients to collaboratively train powerful graph neural networks without sharing their private, decentralized graph data. Inherited from generic federated learning, FGL is critically challenged by statistical heterogeneity, where non-IID data distributions across clients can severely impair model performance. A particularly destructive form of this is class imbalance, which causes the global model to become biased towards majority classes and fail at identifying rare but critical events. This issue is exacerbated in FGL, as nodes from a minority class are often surrounded by biased neighborhood information, hindering the learning of expressive embeddings. To grapple with this challenge, we propose GraphFedMIG, a novel FGL framework that reframes the problem as a federated generative data augmentation task. GraphFedMIG employs a hierarchical generative adversarial network where each client trains a local generator to synthesize high-fidelity feature representations. To provide tailored supervision, clients are grouped into clusters, each sharing a dedicated discriminator. Crucially, the framework designs a mutual information-guided mechanism to steer the evolution of these client generators. By calculating each client's unique informational value, this mechanism corrects the local generator parameters, ensuring that subsequent rounds of mutual information-guided generation are focused on producing high-value, minority-class features. We conduct extensive experiments on four real-world datasets, and the results demonstrate the superiority of the proposed GraphFedMIG compared with other baselines.

\end{abstract}

\begin{links}
    \link{Code}{https://github.com/NovaFoxjet/GraphFedMIG}
\end{links}

\section{Introduction}
%\begin{figure}[t]
%\centering
%\includegraphics[width=1\columnwidth]{Main/figures/intro-pre.pdf} 
%\caption{The performance of various GNN models and their GAN versions on Elliptic and Facebook datasets.}
%\label{intro-pre}
%\end{figure}
\begin{figure}[t]
\centering
\includegraphics[width=1\columnwidth]{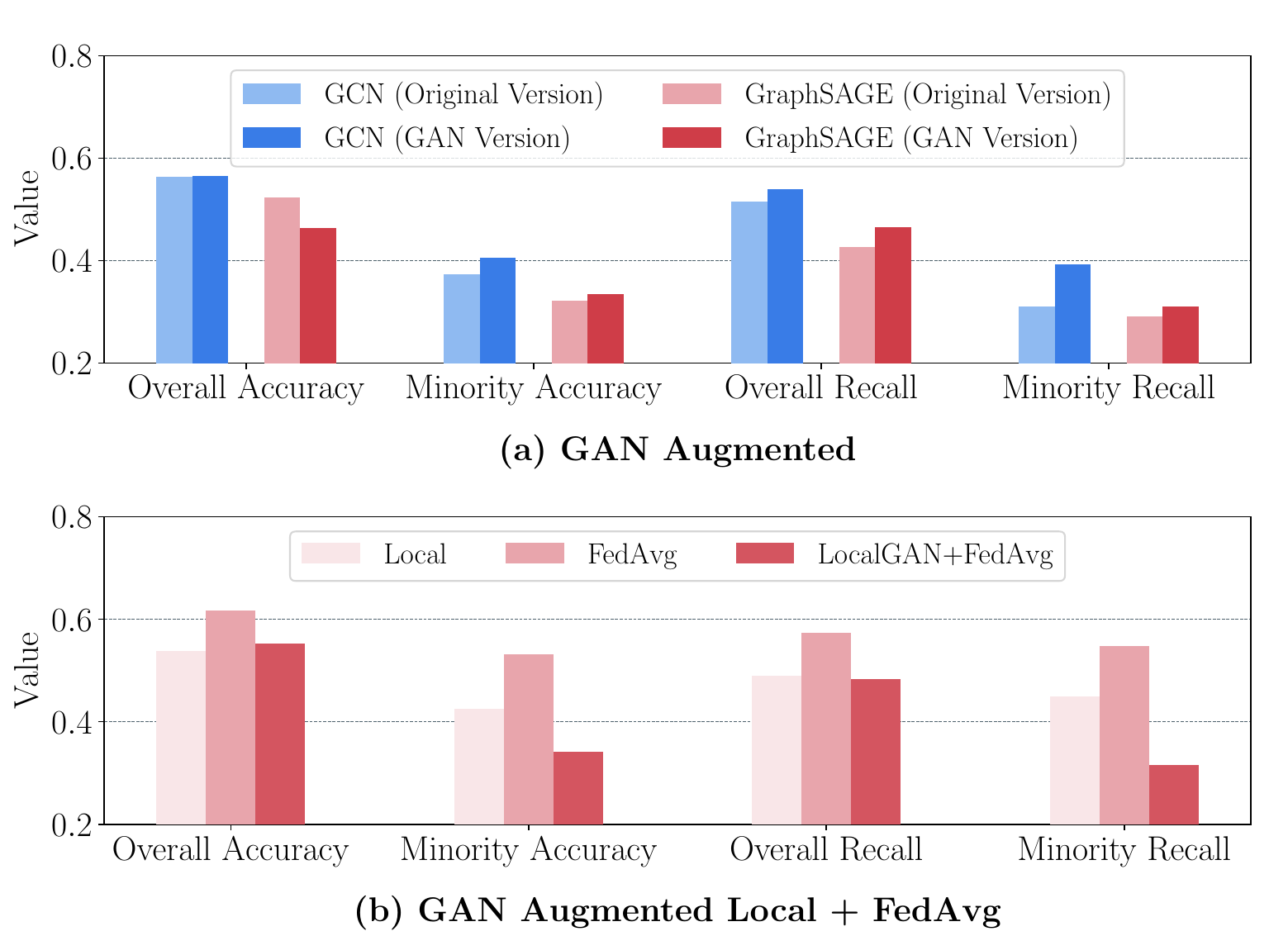} 
\caption{Preliminary experiments on Facebook dataset. (a) Performance comparison of GCN and GraphSAGE, showing their original versions versus versions augmented with a GAN. (b) Performance comparison of three federated strategies: local-only training (Local), standard federated averaging (FedAvg), and LocalGAN+FedAvg.}
\label{intro-pre2}
\end{figure}
Federated graph learning (FGL) represents a critical frontier, uniting the privacy-preserving principles of federated learning (FL) with the expressive power of graph neural networks (GNNs), such as graph convolutional networks (GCN) \cite{kipf2016semi} and GraphSAGE \cite{hamilton2017inductive}. This synthesis allows multiple institutions to collaboratively train sophisticated models on their respective graph-structured data without centralizing sensitive information \citep{liu2024federated, pan2024towards}. Consequently, FGL offers a transformative approach for high-stakes domains like finance \cite{tang2024credit}, healthcare \cite{tang2024personalized}, and recommender systems \cite{zhang2024gpfedrec}, enabling the development of more robust models while strictly adhering to privacy constraints.

However, the practical application of FGL is critically hindered by statistical heterogeneity, where data is not independent and identically distributed (i.e., non-IID) across clients \citep{liu2024federated,yang2024federated}. Among its various forms, class imbalance is a particularly destructive and common failure point \cite{ma2025class}. In real-world FGL scenarios, such as fraud detection, legitimate transactions (majority class) greatly outnumber fraudulent ones (minority class). Standard FGL models trained on such data tend to be biased towards the majority class. This issue is intensified in FGL, as nodes from a minority class are often surrounded by biased neighborhood information, hindering the learning of meaningful embeddings. Consequently, models may achieve high overall accuracy but perform poorly on the minority class, failing to detect rare but crucial events. This limitation undermines FGL’s utility in high-stakes scenarios, where metrics like minority-class accuracy and recall are essential.

Existing FGL methods \citep{shen2022agnostic,zhang2023fedcp} often attempt to mitigate class imbalance through model-centric strategies like re-weighting or modifying the aggregation process. These approaches, however, only compensate for the lack of minority data rather than addressing the root cause: data scarcity. This limitation motivates a paradigm shift from model-level compensation to data-level augmentation. \textit{We hypothesize that a generative approach, which can synthesize high-quality, representative features for underrepresented classes, could directly solve the data scarcity problem at its source.}

%To validate this, we conduct a preliminary experiment comparing standard GNNs (GCN \cite{kipf2016semi} and GraphSAGE \cite{hamilton2017inductive}) against variants augmented with a generative adversarial network (GAN) \cite{gui2021review} on the Elliptic and Facebook datasets.
%As shown in Figure \ref{intro-pre}, GAN-based augmentation acts as a powerful, model-agnostic enhancer, consistently and significantly improving minority-class performance across different GNN backbones and datasets. This improvement, however, can come at the expense of overall performance, as evidenced by the decrease in overall accuracy on the Facebook dataset. These findings affirm that a generative solution is a promising direction for combating class imbalance in FGL, but its direct application is non-trivial. 

To validate our hypothesis, we conduct preliminary experiments that eveal a critical dichotomy in using generative models for class imbalance. First, as shown in Figure \ref{intro-pre2}(a), generative adversarial network (GAN)-based augmentation applied to a single client on the Facebook dataset significantly improves minority-class performance, affirming its potential. However, this promise is not easily realized in a federated context. Our second experiment, illustrated in Figure \ref{intro-pre2}(b), shows that a naive federated strategy combining local GANs with standard FedAvg (LocalGAN+FedAvg) results in a catastrophic performance degradation for the minority class. We attribute this failure to mode collapse, a classic GAN instability issue exacerbated by federated aggregation. The naive averaging of generators destabilizes the adversarial training process, causing the global model to lose the very minority class patterns it is meant to learn. These findings show that while generative augmentation is promising, its direct application in FGL is unstable and counterproductive. Therefore, the central challenge is \textit{designing a federated GAN framework that can remain stable and effective when faced with the inherent instability of adversarial training, a problem exacerbated by class imbalance.}

To this end, we propose \textbf{GraphFedMIG}, a novel framework for \textbf{Graph} \textbf{Fed}erated learning that leverages \textbf{M}utual \textbf{I}nformation-guided \textbf{G}eneration. 
To address the core challenge of applying generative models in federated environments, GraphFedMIG introduces a synergistic, two-part strategy. First, to mitigate minority data scarcity at the source, we employ a hierarchical GAN architecture where each client trains a local generator, with cluster-specific discriminators providing tailored supervision. Second, to ensure stability and effectiveness in the federated generation process, we introduce a mutual information-guided mechanism. Inspired by information theory, this process calculates each client's unique informational value to apply a corrective update to their local generator. This ensures clients with rare, critical data exert greater influence on the system's generative capabilities and directly counteracts aggregation bias. As a result, GraphFedMIG successfully harnesses generative augmentation for class imbalance while maintaining the stability required for complex federated environments.
Our main contributions are summarized as follows.
\begin{itemize}
\item We propose GraphFedMIG, a novel FGL paradigm that treats class imbalance as a generative data augmentation task. Its hierarchical GAN architecture directly mitigates the scarcity of minority data at the source.
\item We design an information-theoretic update scheme to govern the federated generation process. This scheme corrects each client generator based on its unique informational value, thereby serving as a principled regularizer that mitigates systemic bias against minority class data while stabilizing adversarial training.
\item We conduct extensive experiments on multiple real-world datasets. The results demonstrate that GraphFedMIG significantly outperforms state-of-the-art methods.
\end{itemize}
\section{Related Work}
\subsection{Federated Learning}

To address the challenge of statistically heterogeneous data, personalized federated learning (PFL) \cite{fallah2020personalized} has emerged as a key paradigm, aiming to tailor the models to the unique local data distribution of each client. 
A prominent strategy is parameter adaptation, where methods like APFL \cite{deng2020adaptive} generate personalized models by learning to linearly combine the shared global model with each client's own locally trained model. 
Regularization is also employed to guide local training. For instance, \cite{shoham2019overcoming} uses the Fisher information matrix \cite{fisher1970statistical} to penalize large deviations from the global model. 
Another category enhances inter-client collaboration, either by clustering similar clients for group-specific models \citep{xie2021federated,marfoq2021federated,briggs2020federated} or by enabling direct information exchange through secondary communication channels \citep{mori2022continual}.

When heterogeneity manifests as class imbalance, these model-centric methods can only compensate for data scarcity through parameter-space strategies. Some works propose data-level solutions, such as local oversampling or down-sampling \citep{duan2020self,xiao2023triplets}. However, these methods often overlook the unique structural properties of graph data and, critically, do not address the aggregation bias that systematically devalues minority information at the server.

\subsection{Federated Graph Learning}
Federated graph learning (FGL) extends FL to distributed graphs, incurring a new challenge of handling heterogeneous graph topologies across clients. 
Most existing works address this by statically leveraging graph structure, primarily through augmenting structural representations \cite{huang2024federated,wang2024graphproxy}. For instance, FedSAGE+ \cite{zhang2021subgraph} integrates a generative model to create synthetic embeddings during the message passing phase, handling incomplete graph structures. FedStar \cite{tan2023federated} generates structural embeddings for nodes by performing random walks, while FedSpray \cite{fu2024federated} proposes ``structural proxies" that encapsulate both intrinsic node features and local neighborhood information. 
Other approaches mainly tackle the problem from a privacy or prototype perspective. FedSGC \cite{cheung2021fedsgc} employs homomorphic encryption to enable the private exchange of adjacency matrices and node features. 
Methods \cite{huang2023rethinking} based on federated prototype learning extract salient features and structural patterns into compact prototypes that subsequently guide local model training.

%These methods typically treat the graph structure as static or rely on a fixed encoder, which prevents the structural information from co-adapting with the model's evolving representations. Our work differs by dynamically generating feature representations that evolve with the training process.
These methods are limited by their reliance on static structural encoders and, more critically, lack a stable generative framework to address systemic class imbalance. Our work tackles these challenges by dynamically generating rich and stable features that empower a more effective, minority-aware aggregation.
\section{Preliminary}

\subsection{Graph Neural Network}
We define an undirected graph as $\mathcal{G} = (\mathcal{V}, \mathcal{E}, \mathbf{X})$, where $\mathcal{V} = \{v_1, \dots, v_n\}$ is the node set, $\mathcal{E}$ is the edge set, and $\mathbf{X} \in \mathbb{R}^{|\mathcal{V}| \times d_V}$ is the node feature matrix. The neighbors of a node $v_i$ are denoted by $\mathcal{N}(v_i)$.
A standard $L$-layer GNN iteratively updates the embedding vector $\mathbf{h}_{i}^{l}$ for each node $v_i$ at the $l$-th layer using a parameterized message-passing function $f^{l}$:

\begin{equation}
\label{eq:gnn_message_passing}
\mathbf{h}_{i}^{l} = f^{l}(\mathbf{h}_{i}^{l-1}, \{\mathbf{h}_{j}^{l-1} : v_j \in \mathcal{N}(v_i)\}; \theta^{l}),
\end{equation}
where $\theta^{l}$ are learnable parameters and the initial embedding $\mathbf{h}_{i}^{0}$ is the raw feature vector $\mathbf{x}_{i}$.

After $L$ layers, the final embedding $\mathbf{h}_{i}^{L}$ is used for node classification, typically via a softmax function: $\hat{\mathbf{y}}_i = \text{softmax}(\mathbf{h}_{i}^{L})$ \cite{hamilton2017inductive}.

\subsection{Federated Graph Learning}
%Consider a FL system with $M$ clients, where each client $m$ holds a private dataset $D_m=\left\{\left(x_{j}^{m}, y_{j}^{m}\right)\right\}_{j=1}^{N_m}$.
%The global optimization objective of FL is to solve:
%\begin{equation}
%  \min_{\theta_1,\dots,\theta_M} \sum_{m=1}^M \frac{N_m}{N} \mathcal{L}_m(\theta_m; \mathcal{D}_m),\label{fl-objective}
%\end{equation}
%where $N = \sum_{m=1}^{M} N_m$ is the total sample size, $\mathcal{L}_m$ is the local loss for client $m$, and $\theta_m$ represents its model parameters.
%Classical FL, exemplified by FedAvg, enforces a single global model ($\theta_m = \theta$ for all $m$) by periodically averaging local parameters: $\theta = \sum_{m=1}^{M} \frac{N_m}{N}\theta_m$. 
%However, this uniform approach degrades performance in non-IID data \cite{fu2022federated}. To mitigate this, PFL methods allow each model $\theta_m$ to be better tailored to its local data.

%In our FGL setting, $\theta_m$ and $D_m$, denote the parameters of client $m$'s GNN model and its local graph dataset, respectively. For our node-level classification task, the loss $\mathcal{L}_m$ in Eq. (\ref{fl-objective}) is instantiated as the standard cross-entropy loss.

In a FGL system with $M$ clients, each client $m$ holds a private local graph dataset $\mathcal{D}_m$ consisting of $\mathcal{G}_m = (\mathcal{V}_m, \mathcal{E}_m, \mathbf{X}_m)$ and the corresponding node labels $\mathbf{Y}_m$. Let $N_m$ be the number of nodes on client $m$.
The classical approach, FedAvg \cite{mcmahan2017communication}, aims to learn a single global model with parameters $\theta$ by minimizing the weighted average of local losses:
\begin{equation}
\min_{\theta} \sum_{m=1}^M \frac{N_m}{N} \mathcal{L}_m(\theta; \mathcal{D}_m),
\label{eq:fedavg_objective}
\end{equation}
where $N = \sum_{m=1}^{M} N_m$ is the total number of nodes. This is achieved by having clients train the model on their local data and periodically averaging their parameters at a central server: $\theta = \sum_{m=1}^{M} \frac{N_m}{N}\theta_m$.
However, this uniform approach degrades performance in non-IID data \cite{fu2022federated}. To overcome this limitation, PFL aims to learn a unique model $\theta_m$ for each client, better tailoring it to the specific characteristics of the client's local data. 
In our node-level classification task, $\mathcal{L}_m$ represents the standard cross-entropy loss computed over the nodes in client $m$'s local graph.
\section{Methodology}

\begin{figure*}[t]
\centering
\includegraphics[width=1\textwidth]{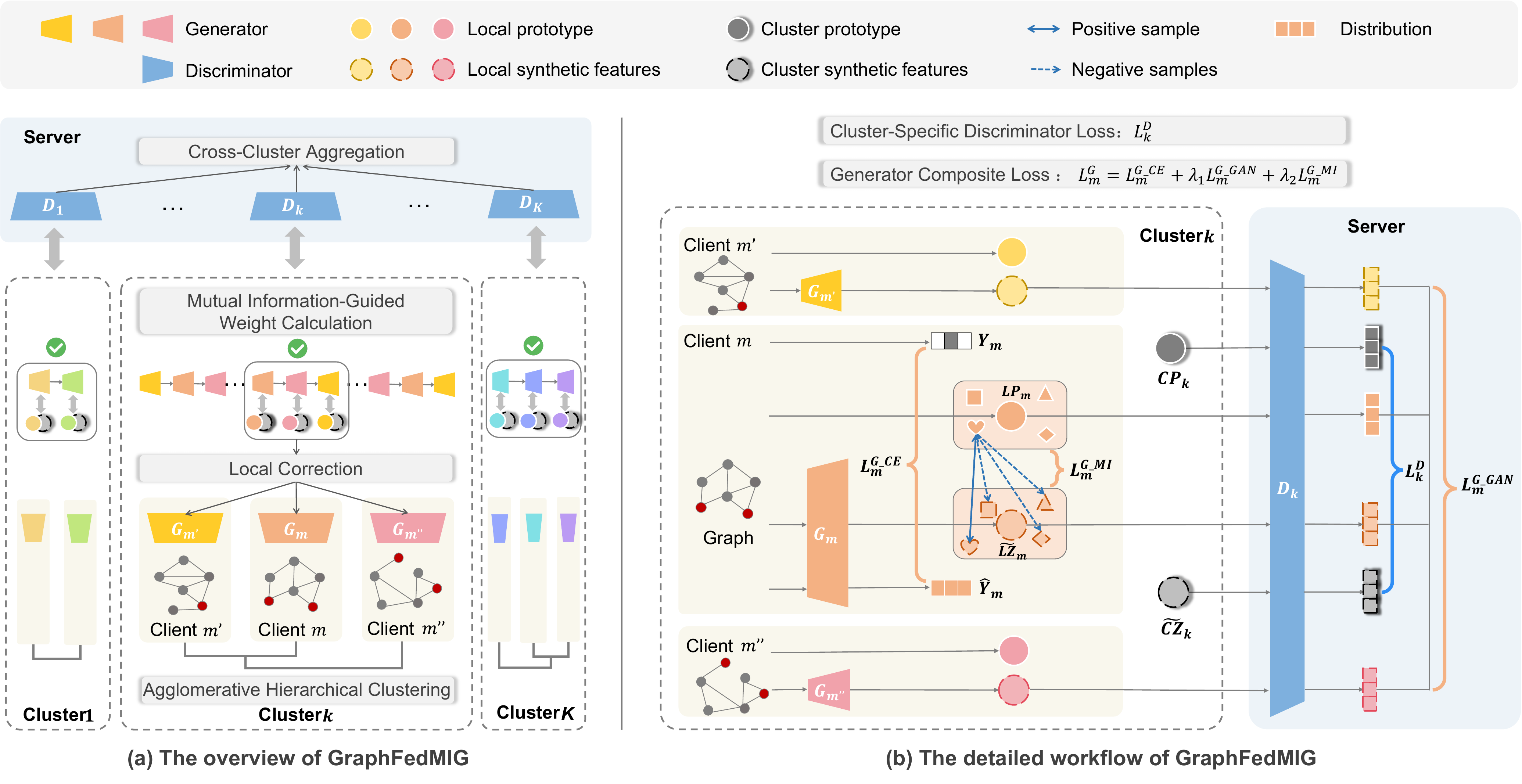} 
\caption{(a) An overview of the proposed GraphFedMIG, a hierarchical GAN framework for FGL. It features client-side generators supervised by cluster-specific discriminators. The framework includes a mutual information-guided method for correcting generator parameters and a cross-cluster aggregation strategy on the server for updating discriminator parameters. (b) An illustration of the detailed workflow and the specific loss functions that drive the model.}
\label{method-framework}
\end{figure*}

In this section, we propose GraphFedMIG, a novel FGL paradigm via generative data augmentation pathway.
As illustrated in Figure~\ref{method-framework}, we first introduce a hierarchical generative adversarial network to learn client-specific data distributions and synthesize high-fidelity node embeddings in a decentralized, privacy-preserving manner. A composite loss function guides this process by balancing classification accuracy with generative diversity and fidelity to local data. Finally, we propose a mutual information-guided aggregation algorithm that personalizes model fusion by weighting client contributions based on their informational value, thereby mitigating aggregation bias.

\subsection{Hierarchical Generative Adversarial Network}
Previous works in FGL often rely on static structural knowledge, which cannot dynamically adapt during training or adequately represent minority classes.
To address this rigidity, we employ a GAN \cite{goodfellow2020generative} to dynamically generate high-fidelity node embeddings.
However, naively integrating GANs into a federated setting is challenging, as a single global discriminator fails to capture client heterogeneity, while deploying client-specific discriminators is computationally prohibitive.
We resolve this with a hierarchical clustering-based GAN that balances computational cost and personalized supervision. This framework first partitions clients into clusters based on data distribution similarity and then assigns a dedicated, shared discriminator to each cluster.

\textbf{Local Generator for Feature Representation.}
For each client $m$, we construct a generator $G_m (\cdot;\theta_m)$ by augmenting a standard GNN backbone with a terminal linear adaptation layer. This layer maps the GNN's node embeddings to a feature space suitable for the generative task. The generator takes node attributes $\mathbf{x}_i$ as input and produces a synthetic feature vector $\widetilde{\mathbf{lz}}_i$ for each node $v_i$. The set of these vectors for client $m$ is denoted by $\widetilde{\mathcal{LZ}}_m = \{\widetilde{\mathbf{lz}}_i\}_{i=1}^{|\mathcal{V}_m|}$.

\textbf{Client Agglomerative Hierarchical Clustering.}
To effectively group clients with similar data distributions, we perform agglomerative hierarchical clustering across clients \cite{bouguettaya2015efficient}. 
%This process uses class-conditional mean features as a proxy for each client's data distribution. 
To ensure stable and efficient training, we perform a one-time clustering of clients as a pre-processing step. This clustering is based on stable initial representations obtained by pre-training a GNN on each client's local data.
Specifically, for each client $m$, we compute the mean feature vector for every class $h\in\{1,\dots,H\}$ based on its locally generated features: ${\widetilde{\mathbf{lz}}}_m^h=\frac{\sum_{i \in \mathcal{V}_m^{h}} \widetilde{\mathbf{lz}}_i}{|\mathcal{V}_m^{h}|}$, where $\mathcal{V}_m^{h}$ is the set of nodes on client $m$ belonging to class $h$. The clustering algorithm proceeds iteratively: at each step, the two clusters, $c_a$ and $c_b$, with the highest similarity measure $S(\cdot,\cdot)$ between their representative mean vectors are merged. The mean vector for the new cluster is updated by averaging the vectors of its constituents: $\widetilde{\mathbf{cz}}^h_{c_{new}}=({\widetilde{\mathbf{cz}}_{c_a}^h+\widetilde{\mathbf{cz}}_{c_b}^h})/{2}$. This process continues until the highest similarity between any two clusters falls below a predefined threshold $T$, resulting in a final set of $K$ distinct client clusters, $\mathcal{C} = \{c_1, c_2,\cdots,c_K\}$.

\textbf{Privacy-Preserving Feature Prototypes.}
To train the generators without direct access to real data, we introduce feature prototypes as a privacy-preserving proxy for the true data distribution \citep{tan2022fedproto,mu2023fedproc}. First, each client $m$ computes its local prototype for each class $h$ as the centroid of its real feature vectors:
\begin{equation}
\mathbf{lp}^h_m = \frac{1}{|\mathcal{D}_m^{h}|} \sum_{i \in \mathcal{D}_m^{h}} \mathbf{lz}_i,
\label{eq:my_label}
\end{equation}
where $\mathcal{D}_m^{h}$ is the set of real data samples on client $m$ for class $h$ and $\mathbf{lz}_i$ is the corresponding real feature vector.
Next, for each cluster $c_k\in\mathcal{C}$, we create an aggregated cluster prototype $\mathbf{cp}_k^h$ by calculating the weighted average of the local prototypes from all clients within that cluster:
\begin{equation}
    \mathbf{cp}^h_k = \frac{\sum_{m\in c_k}|\mathcal{D}_m^{h}| \cdot \mathbf{lp}^h_m}{\sum_{m\in c_k}|\mathcal{D}_m^{h}|}.
\end{equation}

The set of these aggregated prototypes, $\mathcal{CP}_k=\{\mathbf{cp}^1_k,\dots,\mathbf{cp}^H_k\}$, encapsulates the representative data distribution for cluster $c_k$ and serves as the ``real" data target for the cluster's discriminator.

\textbf{Cluster-Specific Discriminator.}
Finally, we define a shared discriminator $D_k(\cdot;\phi_k)$ for each cluster $c_k$. Its objective is to assess whether a synthetic feature aligns with the cluster's prototype distribution. 
The discriminator loss is formulated as the categorical cross-entropy between the probability distribution it predicts for synthetic features and the distribution it predicts for the true cluster prototypes:
\begin{equation}
    \mathcal{L}^D_k = \text{CE}(D_k(\widetilde{\mathcal{CZ}}_k),D_k(\mathcal{CP}_k)),
    \label{eq:placeholder}
\end{equation}
where $\widetilde{\mathcal{CZ}}_k=\{\widetilde{\mathbf{cz}}_k^1,\dots,\widetilde{\mathbf{cz}}_k^H \}$ is the set of synthetic features generated by clients in cluster $c_k$. In this objective, the discriminator's output for the true prototypes, $D_k(\mathcal{CP}_k)$ serves as a dynamic, ``soft" target distribution. This creates a powerful feedback loop: as client GNNs improve, their feature embeddings become more refined, leading to higher-quality prototypes. This allows the generative target to co-evolve with the models, a significant advantage methods relying on fixed proxy data.

\subsection{Generator Optimization via Composite Loss}
%The generator $G_m$ is optimized to serve multiple, sometimes competing, objectives: maintaining its primary utility as a classifier, producing diverse synthetic samples to train the discriminator effectively, and ensuring the synthesized knowledge remains faithful to the client's local data distribution. To balance these demands, we formulate a composite loss function with three components.
The generator $G_m$ is optimized via a composite loss function balancing three objectives: classification accuracy, generative diversity, and fidelity to the local data distribution.

\textbf{Classification Loss.}
To ensure the generator's GNN backbone maintains its primary classification utility, we apply a standard cross-entropy loss. This loss is computed on client $m$'s local labeled data $(\mathcal{D}_m,\mathcal{Y}_m)$.
\begin{equation}
\mathcal{L}^{G\_{CE}}_m = \text{CE}(G_m(\mathcal{D}_m), \mathcal{Y}_m),
\end{equation}
where $G_m(\mathcal{D}_m)$ denotes the classification output from the GNN backbone for the real data on client $m$.

\textbf{Adversarial and Diversity Loss.}
To encourage the generation of diverse yet relevant samples, we establish a ``cooperative competition" among the generators within a cluster. 
Each generator $G_m$ is incentivized not only to fool the discriminator but also to align its output distribution with the average distribution produced by its peers in the cluster. This promotes a collaborative exploration of the feature space to collectively approximate the target prototype distribution. This structured diversification uses the following Kullback-Leibler (KL) divergence objective \cite{kullback1951information} (see \textbf{Appendix A} for detailed derivation).
{\small
\begin{multline}
    \mathcal{L}^{G\_{GAN}}_m = \text{KL}\left(P(y|m) \| \frac{P(y|m)+\sum_{m'\in c_k} P(y|G_{m'})}{2}\right) \\
    + H \text{KL} \left(\sum_{m'\in c_k} P(y|G_{m'}) \| \frac{P(y|m)+\sum_{m'\in c_k} P(y|G_{m'})}{2}\right) \\
    - (H+1) \log (H+1) + H \log (H),
\end{multline}
}where $P(y|m)$ represents client $m$'s true class distribution (approximated by its local prototype), $P(y|G_{m'})$ is the class distribution produced by generator $G_{m'}$.

\textbf{Mutual Information Fidelity Loss.}
To counteract distributional drift under data heterogeneity, we introduce a mutual information loss that serves as a stabilizing anchor. This loss maximizes the mutual information between a client's local, real feature prototypes $\mathbf{lp}_m^h$ and the synthetic features $\widetilde{\mathbf{cz}}_k^h$ generated for the corresponding class within its cluster $c_k$. This ensures the generative process remains grounded in the client's specific data characteristics. The objective is implemented using the InfoNCE loss \citep{velivckovic2018deep} , a tractable lower bound on mutual information:
\begin{equation}
\mathcal{L}^{G\_{MI}}_{m} = - \mathbb{E} \left[ \log \frac{\exp(f(\mathbf{lp}_m^h)^{\top} f(\widetilde{\mathbf{cz}}_k^h))}{\sum_{j=1}^{H} \exp(f(\mathbf{lp}_m^h)^{\top} f(\widetilde{\mathbf{cz}}_k^j))} \right].
\end{equation}
Here, for a client $m\in c_k$, the positive pair consists of its real prototype $\mathbf{lp}_m^h$ and a synthetic feature $\widetilde{\mathbf{cz}}_k^h$ of the same class $h$. The negative pairs consist of $\mathbf{lp}_m^h$ and synthetic features $\widetilde{\mathbf{cz}}_k^j$ from all other classes $j\neq h$.
The function $f(\cdot)$ is a projection head, a small neural network, that maps features into a dedicated space optimized for contrastive comparison.

\textbf{Final Composite Loss.}
The final objective for generator $G_m$ is a weighted sum of the three components, balanced by hyperparameters $\lambda_1$ and $\lambda_2$.
\begin{equation}
    \mathcal{L}^G_m = \mathcal{L}^{G\_{CE}}_m + \lambda_1 \mathcal{L}^{G\_{GAN}}_{m} + \lambda_2 \mathcal{L}^{G\_{MI}}_{m}.
    \label{eq:placeholder}
\end{equation}

\subsection{Mutual Information-Guided Model Aggregation} 

The final stage of each communication round involves updating the client-side generators and server-side discriminators. GraphFedMIG departs from traditional federated averaging by employing a hybrid approach: client generators undergo a personalized, mutual information-guided local correction, while the cluster discriminators are aggregated globally to facilitate knowledge sharing across clusters.

\textbf{Mutual Information-Guided Weight Calculation.}
To derive a principled weight $\mathbf{W}_m$ for our local correction mechanism, we first model the clients' interaction within a cluster $c_k$, through a theoretical construct: a virtual cluster generator $\mathbf{\hat{y}}_k$. This construct is designed to be invariant to client ordering (client-level permutation invariance). We formally define this as:
\begin{equation}
  \mathbf{\hat{y}}_k = \sigma\left( \sum_{m \in c_k} \mathbf{W}_m G_m  \boldsymbol{\pi}^{\top} \right),
\label{eq:virtual-generator}
\end{equation}
where $\mathbf{\hat{y}}_k$ conceptually synthesizes the knowledge from all generators $G_m$ in the cluster. The term $\mathbf{W}_m$ is the client-specific contribution weight, and $\boldsymbol{\pi}$ is a permutation matrix that ensures the combination is robust to the client ordering. 

With this permutation-invariant structure established, we can formulate an objective to find the optimal contribution weight $\mathbf{W}_m$ for each client. We posit that a client's influence should be proportional to its informational value. This is framed as an optimization problem that maximizes the information-weighted contribution of each client, linking its weight $\mathbf{W}_m$ to its mutual information $\text{MI}_m$:
\begin{equation}
  \max_{\mathbf{W}_m} \sum_{m \in c_k} \mathbf{W}_m \text{MI}_m \boldsymbol{\pi}^\top \mathbf{W}_m.
  \label{eq:aggregation-weight}
\end{equation}

The mutual information $\text{MI}_m$ quantifies the statistical dependency between a client's local data distribution and the aggregated distribution of its cluster $c_k$. 
%It is estimated by comparing the client's posterior class distribution $P(y|m)$ (approximated by its local prototype) with the cluster's average posterior $P(y|c_k)$ (approximated using the cluster-generated prototype). 
This dependency is measured as the Jensen-Shannon (JS) divergence between the two distributions, which is estimated using a robust, non-binning nearest-neighbor method \cite{ross2014mutual}. The client's posterior class distribution $P(y|m)$ (approximated by its local prototype) and the cluster's average posterior $P(y|c_k)$ (approximated using the cluster-generated prototype) serve as the inputs for this estimation.
A high $\text{MI}_m$ value signifies that the client contributes unique class information, especially for classes underrepresented within the cluster, justifying a higher weight.

\textbf{Local Correction for Generators.}
To enhance personalization, we replace standard aggregation with a local parameter correction. Each client $m$ adaptively re-scales its own generator parameters $\theta_m$ using its calculated weight $\mathbf{W}_m$:
\begin{equation}
    \tilde{\theta}_m = \mathbf{W}_m \theta_m.
    \label{eq:generator-update}
\end{equation}

This allows each generator to evolve along a personalized trajectory modulated by its unique contribution, rather than being pulled towards a generic average.

\textbf{Cross-Cluster Aggregation for Discriminators.}
While generators are updated locally, the $K$ cluster-specific discriminators are aggregated at the server to share knowledge across clusters. This prevents discriminators from becoming overly specialized and allows them to maintain a global perspective. The aggregation is a weighted average based on the relative size of each cluster:
\begin{equation}
     \phi_{global} = \sum_{k = 1}^{K} \frac{|c_k|}{M} \phi_k.
\end{equation}

The resulting $\phi_{global}$ is used to initialize the cluster discriminators for the next round.
\section{Experiment}
In this section, we describe our experimental settings and subsequently present the results from extensive experiments designed to evaluate the effectiveness of GraphFedMIG. 
%The code is available at https://anonymous.4open.science/r/GraphFedMIG.

\subsection{Experimental Setup}
\begin{table*}[ht]
    \centering
    \small
    \begin{tabular}{c|c|c|cccccc}
    \toprule
    Datasets & \multicolumn{2}{c|}{Metrics} &  Local &  FedAvg &  FL+HC &  FedSpray &  \textbf{GraphFedMIG} &   Improv. \\
    \midrule
    \multirow{4}{*}{Elliptic}  & \multirow{2}{*}{Accuracy}  &    Overall  &  0.819  &   0.837  &  \underline{0.871}  &     0.862  &        \textbf{0.894}  &  2.64\% \\
     &  &   Minority  &  0.416  &   0.463  &  \underline{0.695}  &     0.530  &        \textbf{0.740}  &  6.47\% \\
    \cmidrule{2-9}
     & \multirow{2}{*}{Recall}  &    Overall  &  0.686  &   0.715  &  \underline{0.816}  &     0.757  &        \textbf{0.849}  &  4.04\% \\
     &  &   Minority  &  0.423  &   0.457  &  \underline{0.696}  &     0.520  &        \textbf{0.733}  &  5.31\% \\
    \midrule
    \multirow{4}{*}{Twitch}  & \multirow{2}{*}{Accuracy}  &    Overall  &  0.587  &   0.568  &  0.573  &     \textbf{0.611}  &        \underline{0.597}  & -2.29\% \\
     &  &   Minority  &  \underline{0.496}  &   0.487  &  0.473  &     0.442  &        \textbf{0.515}  &  3.83\% \\
    \cmidrule{2-9}
     & \multirow{2}{*}{Recall}  &    Overall  &  0.564  &   0.550  &  0.543  &     \underline{0.575}  &        \textbf{0.582}  &  1.22\% \\
     &  &   Minority  &  0.485  &   \underline{0.493}  &  0.456  &     0.437  &        \textbf{0.520}  &  5.48\% \\
    \midrule
    \multirow{4}{*}{Facebook}  & \multirow{2}{*}{Accuracy}  &    Overall  &  0.538  &   0.617  &  \underline{0.683}  &     0.642  &        \textbf{0.703}  &  2.93\% \\
     &  &   Minority  &  0.425  &   0.532  &  0.481  &     \underline{0.559}  &        \textbf{0.637}  &  13.95\% \\
    \cmidrule{2-9}
     & \multirow{2}{*}{Recall}  &    Overall  &  0.489  &   0.574  &  0.523  &     \underline{0.597}  &        \textbf{0.655}  &  9.72\% \\
     &  &   Minority  &  0.449  &   0.548  &  0.493  &     \underline{0.573}  &        \textbf{0.648}  &  13.09\% \\
    \midrule
    \multirow{4}{*}{Actor}  & \multirow{2}{*}{Accuracy}  &    Overall  &  0.271  &   \underline{0.317}  &  0.312  &     0.314  &        \textbf{0.327}  &  3.15\% \\
     &  &   Minority  &  0.257  &   \underline{0.264}  &  0.224  &     0.252  &        \textbf{0.297}  &  12.50\% \\
    \cmidrule{2-9}
     & \multirow{2}{*}{Recall}  &    Overall  &  0.253  &   0.290  &  \underline{0.306}  &     0.258  &        \textbf{0.319}  &  4.25\% \\
     &  &   Minority  &  0.259  &   \underline{0.269}  &  0.227  &     0.254  &        \textbf{0.297}  &  10.41\% \\
    \bottomrule
    \end{tabular}
    \caption{Performance of GraphFedMIG and other baselines over four datasets. The best results are bold and the best baseline results are underlined. \textit{Improv.} indicates the performance improvement over the best baseline, respectively.}\label{exp-overall-performance}
\end{table*}

\subsubsection{Datasets.} We evaluate GraphFedMIG on four public, class-imbalanced graph datasets from diverse domains: Elliptic (financial risk), Twitch (violation detection), Facebook (social recommendation), and Actor (affiliation network). See \textbf{Appendix B} for further details.

\subsubsection{Evaluation Metrics.} Following prior work \citep{zhang2023survey,li2024aligning}, we measure performance using overall accuracy, minority accuracy, overall recall, and minority recall. Further details are in \textbf{Appendix C}.

\subsubsection{Baselines.} We compare GraphFedMIG with four baselines, all using GraphSAGE \cite{hamilton2017inductive} as the base GNN: Local (local training), FedAvg \cite{mcmahan2017communication}, FL+HC \cite{briggs2020federated} (clustered FL), and the state-of-the-art FedSpray \cite{fu2024federated}. See \textbf{Appendix D} for baseline details.

\subsubsection{Implementation Details.} We implement the methods based on Pytorch framework and the hyperparameter configuration is summarized in \textbf{Appendix E}.

\subsection{Overall Performance Comparison}
As shown in Table \ref{exp-overall-performance}, we present a performance comparison of node classification on four datasets.
First, we examine the overall accuracy on all test nodes. GraphFedMIG achieves the highest performance on three of the four datasets, reaching 89.4\% on Elliptic, 70.3\% on Facebook, and 32.7\% on Actor. While FedSpray slightly outperforms it on the Twitch dataset, GraphFedMIG consistently surpasses the other baselines by a significant margin.

The analysis then shifts to the performance for the minority class, which is the main target of improvement by our method. In these areas, GraphFedMIG demonstrates a significant advantage across all datasets. On the Facebook dataset, for instance, GraphFedMIG achieves a minority recall of 64.8\%, representing a 13.09\% improvement over the second-best baseline. This trend of superiority is consistent across the Twitch, Facebook, and Actor datasets. Furthermore, GraphFedMIG secures the highest overall recall in all scenarios, underscoring its superior ability to correctly identify positive examples.

GraphFedMIG's superior performance stems from its unique design, which overcomes baseline limitations. Methods like FL+HC falter as their clustering can obscure minority features, while FedSpray uses a limited structural proxy that fails to capture rich graph features. In contrast, GraphFedMIG's generator extracts comprehensive local features, and its mutual information-based weighting prioritizes feature quality over data quantity. 
%This ensures effective learning of minority class characteristics, leading to state-of-the-art performance.

\begin{table*}[ht]
    \centering
    \small
    \begin{tabular}{cccccccc}
    \toprule
\multirow{2}{*}{HC} & \multirow{2}{*}{GAN} & \multirow{2}{*}{MI loss} & \multirow{2}{*}{MIGMA} & \multicolumn{2}{c}{Accuracy} & \multicolumn{2}{c}{Recall} \\
\cmidrule(lr){5-6} \cmidrule(lr){7-8}
 & & & & Overall & Minority & Overall & Minority \\
\midrule
 \checkmark &     &         &       &   0.683 &    0.481 &   0.523 &    0.493 \\
 \checkmark &  \checkmark &         &       &   0.574 &    0.487 &   0.528 &    0.501 \\
 \checkmark &   \checkmark &       \checkmark &       &   0.641 &    0.569 &   0.592 &    0.579 \\
 \checkmark &   \checkmark &       \checkmark &     \checkmark &   \textbf{0.703} &    \textbf{0.637} &   \textbf{0.655} &    \textbf{0.648} \\
\bottomrule
\end{tabular}
    \caption{Ablation study on Facebook.}
    \label{exp-ablation-study-performance}
\end{table*}

\subsection{Ablation Study}
\textbf{Effectiveness of Model Components.} Table \ref{exp-ablation-study-performance} presents an ablation study validating each component of the GraphFedMIG framework. The model consists of four core components: hierarchical clustering (\textbf{HC}), a many-to-one generative adversarial network (\textbf{GAN}), a mutual information loss (\textbf{MI loss}), and a mutual information-guided model aggregation (\textbf{MIGMA}) algorithm. 

The analysis shows a clear progression. The baseline HC model achieved a high overall accuracy of 0.683 but performed poorly on the minority class, with a minority accuracy of only 0.481. The subsequent integration of the GAN and MI loss modules addresses this weakness systematically. Notably, the MI loss achieves a boost, elevating minority accuracy to 0.569 and minority recall to 0.579. This confirms their combined role in generating higher-quality synthetic data and enhancing minority feature representation.

The complete GraphFedMIG model, which includes the MIGMA component, attains optimal results across all metrics. It reaches a peak overall accuracy of 0.703 and a minority accuracy of 0.637, demonstrating the component's effectiveness in refining parameter aggregation. For results on other datasets, see \textbf{Appendix F}.

\begin{figure}[t!]
\centering
\includegraphics[width=\columnwidth]{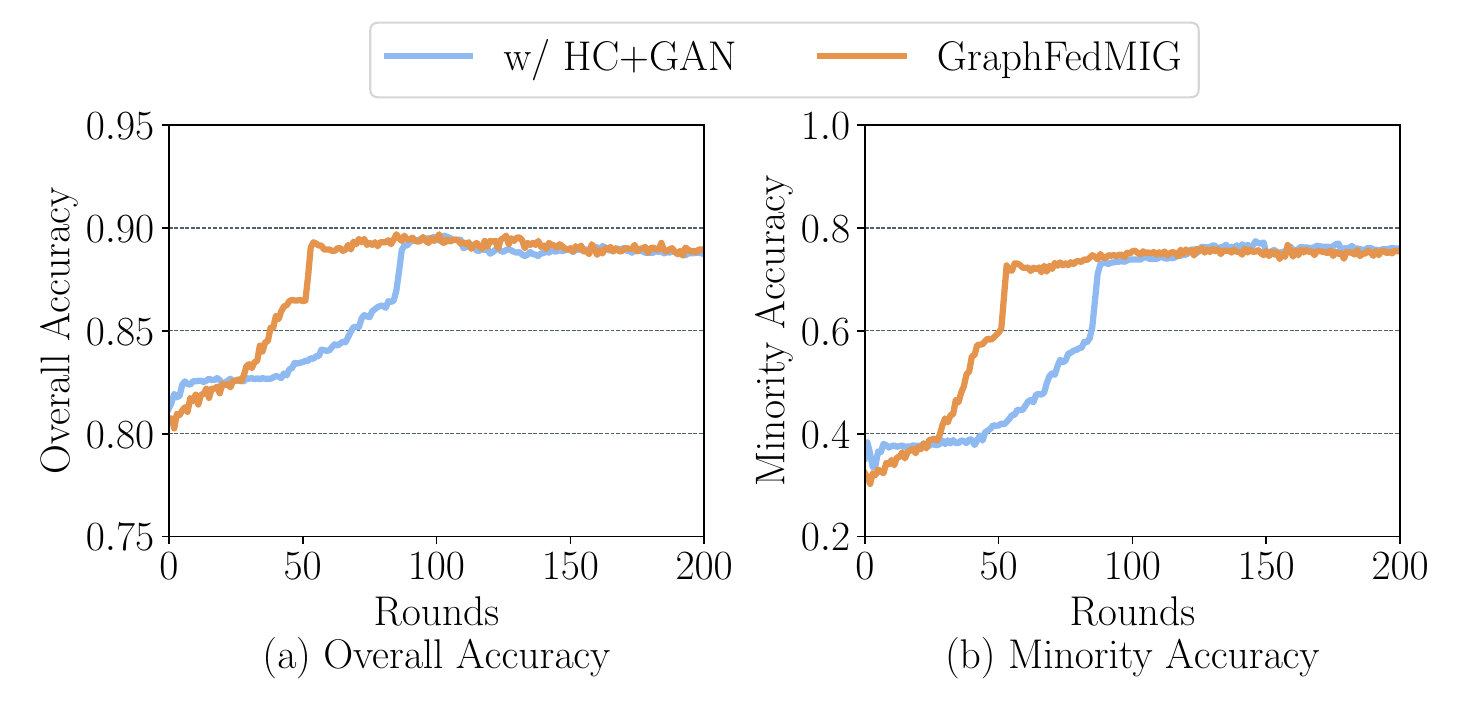} 
\caption{Convergence curves on Elliptic.}\label{exp-ablation-study-convergence}
\end{figure}

\textbf{Analysis of Convergence and Stability.} To evaluate the impact of our proposed components on training stability and convergence speed, we compare the full GraphFedMIG model with an ablation variant, denoted as w/ HC+GAN. This variant incorporates HC and GAN but excludes MI loss and MIGMA. As illustrated in Figure \ref{exp-ablation-study-convergence}, GraphFedMIG demonstrates significantly faster convergence, reaching a high, stable plateau in both overall accuracy and minority accuracy around the 50th communication round. In contrast, the w/ HC+GAN variant exhibits a more volatile and protracted training process, requiring approximately 90 rounds to reach a comparable performance level.

The faster convergence highlights the role of MI loss and MIGMA as powerful regularizers. They enhance generator stability, ensuring high-quality data generation that in turn reduces model fluctuations and improves the overall training efficiency of the GraphFedMIG framework.

\subsection{Analysis of GraphFedMIG}
\begin{figure}[t!]
\centering
\includegraphics[width=\columnwidth]{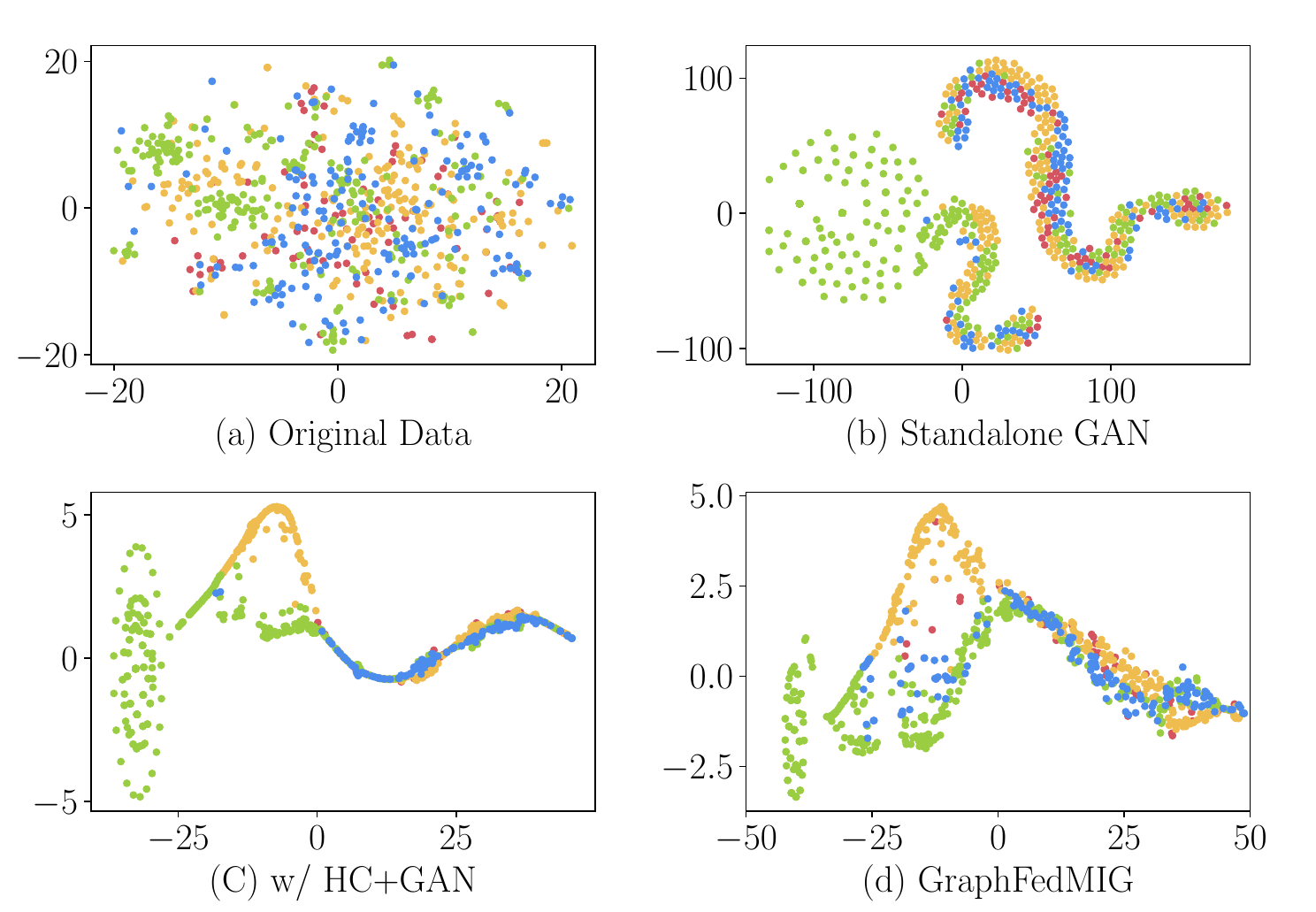} 
\caption{Visualization of data distributions on Facebook.}\label{exp-qualitative-analysis}
\end{figure}

\textbf{Qualitative Analysis.} We present 2D scatter plot visualizations in Figure \ref{exp-qualitative-analysis}, comparing the embeddings of the original data with those generated by the different models. The original data points from the Facebook dataset (a) are highly scattered, exhibiting no discernible patterns or clear class clusters. In contrast, by incorporating graph structural information, the generative models produce data with more defined structures. While the standalone GAN (b) forms distinct non-linear clusters, the w/ HC+GAN ablation variant (c) attempts to further improve class separability for the green and yellow classes but leads to a less stable manifold with significant overlap among other data points. Our proposed GraphFedMIG (d) effectively overcomes these issues. It not only enhances the distinction between all classes, especially the green and yellow ones, but also substantially reduces the overlap among the remaining data points. This comprehensive visualization demonstrates that GraphFedMIG generates higher-quality synthetic data, characterized by more distinct and stable class distributions compared to both the baseline GAN and the w/ HC+GAN variant.

\textbf{More Experimental Results.} Due to the page limit, more experimental results are provided in our technical appendix, including hyperparameter analysis and differential privacy protection, among others.
\section{Conclusion}
In this paper, we address the critical problem of class imbalance in FGL. We identify that the failure of standard FGL methods in such scenarios stems not only from local data scarcity but more fundamentally from a systemic aggregation bias that marginalizes rare but crucial information during model fusion. To overcome this, we propose GraphFedMIG, a novel framework that reframes the challenge as a federated generative data augmentation task.
GraphFedMIG tackles the problem with a dual-pronged strategy. Based on a hierarchical generative adversarial network, high-fidelity features are synthesized to augment the training data. 
Furthermore, GraphFedMIG introduces a mutual information-guided mechanism. This mechanism rectifies aggregation bias by ensuring knowledge from the minority class is preserved and prioritized, which also helps in stabilizing the adversarial training.
Extensive experiments on four real-world datasets validate the superiority of GraphFedMIG.
\bibliography{main}

\begin{thebibliography}{45}
\providecommand{\natexlab}[1]{#1}

\bibitem[{Abadi et~al.(2016)Abadi, Chu, Goodfellow, McMahan, Mironov, Talwar, and Zhang}]{abadi2016deep}
Abadi, M.; Chu, A.; Goodfellow, I.; McMahan, H.~B.; Mironov, I.; Talwar, K.; and Zhang, L. 2016.
\newblock Deep Learning with Differential Privacy.
\newblock In \emph{Proceedings of the ACM SIGSAC Conference on Computer and Communications Security}, 308--318.

\bibitem[{Bouguettaya et~al.(2015)Bouguettaya, Yu, Liu, Zhou, and Song}]{bouguettaya2015efficient}
Bouguettaya, A.; Yu, Q.; Liu, X.; Zhou, X.; and Song, A. 2015.
\newblock Efficient Agglomerative Hierarchical Clustering.
\newblock \emph{Expert Systems with Applications}, 42(5): 2785--2797.

\bibitem[{Briggs, Fan, and Andras(2020)}]{briggs2020federated}
Briggs, C.; Fan, Z.; and Andras, P. 2020.
\newblock Federated Learning with Hierarchical Clustering of Local Updates to Improve Training on Non-{IID} Data.
\newblock In \emph{Proceedings of the International Joint Conference on Neural Networks (IJCNN)}, 1--9.

\bibitem[{Cheung, Dai, and Li(2021)}]{cheung2021fedsgc}
Cheung, T.~H.; Dai, W.; and Li, S. 2021.
\newblock {FedSGC}: Federated Simple Graph Convolution for Node Classification.
\newblock In \emph{Proceedings of the IJCAI Workshops}.

\bibitem[{Deng, Kamani, and Mahdavi(2020)}]{deng2020adaptive}
Deng, Y.; Kamani, M.~M.; and Mahdavi, M. 2020.
\newblock Adaptive Personalized Federated Learning.
\newblock \emph{arXiv preprint arXiv:2003.13461}.

\bibitem[{Duan et~al.(2020)Duan, Liu, Chen, Liu, Tan, and Liang}]{duan2020self}
Duan, M.; Liu, D.; Chen, X.; Liu, R.; Tan, Y.; and Liang, L. 2020.
\newblock Self-Balancing Federated Learning with Global Imbalanced Data in Mobile Systems.
\newblock \emph{IEEE Transactions on Parallel and Distributed Systems}, 32(1): 59--71.

\bibitem[{Fallah, Mokhtari, and Ozdaglar(2020)}]{fallah2020personalized}
Fallah, A.; Mokhtari, A.; and Ozdaglar, A. 2020.
\newblock Personalized Federated Learning: A Meta-Learning Approach.
\newblock \emph{arXiv preprint arXiv:2002.077948}.

\bibitem[{Fisher(1970)}]{fisher1970statistical}
Fisher, R.~A. 1970.
\newblock Statistical Methods for Research Workers.
\newblock In \emph{Breakthroughs in Statistics: Methodology and Distribution}, 66--70. Springer.

\bibitem[{Fu et~al.(2024)Fu, Chen, Zhang, Chen, and Li}]{fu2024federated}
Fu, X.; Chen, Z.; Zhang, B.; Chen, C.; and Li, J. 2024.
\newblock Federated Graph Learning with Structure Proxy Alignment.
\newblock In \emph{Proceedings of the 30th ACM SIGKDD Conference on Knowledge Discovery and Data Mining}, 827--838.

\bibitem[{Fu et~al.(2022)Fu, Zhang, Dong, Chen, and Li}]{fu2022federated}
Fu, X.; Zhang, B.; Dong, Y.; Chen, C.; and Li, J. 2022.
\newblock Federated Graph Machine Learning: A Survey of Concepts, Techniques, and Applications.
\newblock \emph{ACM SIGKDD Explorations Newsletter}, 24(2): 32--47.

\bibitem[{Ghosh et~al.(2018)Ghosh, Kulharia, Namboodiri, Torr, and Dokania}]{ghosh2018multi}
Ghosh, A.; Kulharia, V.; Namboodiri, V.~P.; Torr, P. H.~S.; and Dokania, P.~K. 2018.
\newblock Multi-Agent Diverse Generative Adversarial Networks.
\newblock In \emph{Proceedings of the IEEE Conference on Computer Vision and Pattern Recognition (CVPR)}, 8513--8521.

\bibitem[{Goodfellow et~al.(2020)Goodfellow, Pouget-Abadie, Mirza, Xu, Warde-Farley, Ozair, Courville, and Bengio}]{goodfellow2020generative}
Goodfellow, I.; Pouget-Abadie, J.; Mirza, M.; Xu, B.; Warde-Farley, D.; Ozair, S.; Courville, A.; and Bengio, Y. 2020.
\newblock Generative Adversarial Networks.
\newblock \emph{Communications of the ACM}, 63(11): 139--144.

\bibitem[{Hamilton, Ying, and Leskovec(2017)}]{hamilton2017inductive}
Hamilton, W.; Ying, Z.; and Leskovec, J. 2017.
\newblock Inductive Representation Learning on Large Graphs.
\newblock \emph{Advances in Neural Information Processing Systems}, 30.

\bibitem[{Huang et~al.(2024)Huang, Wan, Ye, and Du}]{huang2024federated}
Huang, W.; Wan, G.; Ye, M.; and Du, B. 2024.
\newblock Federated Graph Semantic and Structural Learning.
\newblock \emph{arXiv preprint arXiv:2406.18937}.

\bibitem[{Huang et~al.(2023)Huang, Ye, Shi, Li, and Du}]{huang2023rethinking}
Huang, W.; Ye, M.; Shi, Z.; Li, H.; and Du, B. 2023.
\newblock Rethinking Federated Learning with Domain Shift: A Prototype View.
\newblock In \emph{Proceedings of the IEEE/CVF Conference on Computer Vision and Pattern Recognition (CVPR)}, 16312--16322.

\bibitem[{Kinga and Ba(2015)}]{kinga2015method}
Kinga, D.; and Ba, J. 2015.
\newblock A Method for Stochastic Optimization.
\newblock In \emph{Proceedings of the International Conference on Learning Representations (ICLR)}.

\bibitem[{Kipf and Welling(2016)}]{kipf2016semi}
Kipf, T.~N.; and Welling, M. 2016.
\newblock Semi-Supervised Classification with Graph Convolutional Networks.
\newblock \emph{arXiv preprint arXiv:1609.02907}.

\bibitem[{Kullback and Leibler(1951)}]{kullback1951information}
Kullback, S.; and Leibler, R.~A. 1951.
\newblock On Information and Sufficiency.
\newblock \emph{The Annals of Mathematical Statistics}, 22(1): 79--86.

\bibitem[{Li, Zhan, and Li(2024)}]{li2024aligning}
Li, L.; Zhan, D.~C.; and Li, X.~C. 2024.
\newblock Aligning Model Outputs for Class Imbalanced Non-{IID} Federated Learning.
\newblock \emph{Machine Learning}, 113(4): 1861--1884.

\bibitem[{Liu et~al.(2024)Liu, Xing, Deng, Li, Guan, and Yu}]{liu2024federated}
Liu, R.; Xing, P.; Deng, Z.; Li, A.; Guan, C.; and Yu, H. 2024.
\newblock Federated Graph Neural Networks: Overview, Techniques, and Challenges.
\newblock \emph{IEEE Transactions on Neural Networks and Learning Systems}.

\bibitem[{Ma et~al.(2025)Ma, Tian, Moniz, and Chawla}]{ma2025class}
Ma, Y.; Tian, Y.; Moniz, N.; and Chawla, N.~V. 2025.
\newblock Class-Imbalanced Learning on Graphs: A Survey.
\newblock \emph{ACM Computing Surveys}, 57(8): 1--16.

\bibitem[{Marfoq et~al.(2021)Marfoq, Neglia, Bellet, Kameni, and Vidal}]{marfoq2021federated}
Marfoq, O.; Neglia, G.; Bellet, A.; Kameni, L.; and Vidal, R. 2021.
\newblock Federated Multi-Task Learning under a Mixture of Distributions.
\newblock \emph{Advances in Neural Information Processing Systems}, 34: 15434--15447.

\bibitem[{McMahan et~al.(2017)McMahan, Moore, Ramage, Hampson, and y~Arcas}]{mcmahan2017communication}
McMahan, B.; Moore, E.; Ramage, D.; Hampson, S.; and y~Arcas, B.~A. 2017.
\newblock Communication-Efficient Learning of Deep Networks from Decentralized Data.
\newblock In \emph{Proceedings of the 20th International Conference on Artificial Intelligence and Statistics (AISTATS)}, 1273--1282.

\bibitem[{Mori, Teranishi, and Furukawa(2022)}]{mori2022continual}
Mori, J.; Teranishi, I.; and Furukawa, R. 2022.
\newblock Continual Horizontal Federated Learning for Heterogeneous Data.
\newblock In \emph{Proceedings of the International Joint Conference on Neural Networks (IJCNN)}, 1--8.

\bibitem[{Mu et~al.(2023)Mu, Shen, Cheng, Geng, Fu, Zhang, and Zhang}]{mu2023fedproc}
Mu, X.; Shen, Y.; Cheng, K.; Geng, X.; Fu, J.; Zhang, T.; and Zhang, Z. 2023.
\newblock {FedProc}: Prototypical Contrastive Federated Learning on Non-{IID} Data.
\newblock \emph{Future Generation Computer Systems}, 143: 93--104.

\bibitem[{Pan et~al.(2024)Pan, Xu, Yu, Yang, Wu, Wang, Chen, and Yang}]{pan2024towards}
Pan, C.; Xu, J.; Yu, Y.; Yang, Z.; Wu, Q.; Wang, C.; Chen, L.; and Yang, Y. 2024.
\newblock Towards Fair Graph Federated Learning via Incentive Mechanisms.
\newblock In \emph{Proceedings of the AAAI Conference on Artificial Intelligence}, 14499--14507.

\bibitem[{Pei et~al.(2020)Pei, Wei, Chang, Lei, and Yang}]{pei2020geom}
Pei, H.; Wei, B.; Chang, K. C.~C.; Lei, Y.; and Yang, B. 2020.
\newblock {Geom-GCN}: Geometric Graph Convolutional Networks.
\newblock \emph{arXiv preprint arXiv:2002.05287}.

\bibitem[{Ross(2014)}]{ross2014mutual}
Ross, B.~C. 2014.
\newblock Mutual Information between Discrete and Continuous Data Sets.
\newblock \emph{PloS One}, 9(2): e87357.

\bibitem[{Rozemberczki, Allen, and Sarkar(2021)}]{rozemberczki2021multi}
Rozemberczki, B.; Allen, C.; and Sarkar, R. 2021.
\newblock Multi-Scale Attributed Node Embedding.
\newblock \emph{Journal of Complex Networks}, 9(2): cnab014.

\bibitem[{Shen et~al.(2022)Shen, Cervino, Hassani, and Ribeiro}]{shen2022agnostic}
Shen, Z.; Cervino, J.; Hassani, H.; and Ribeiro, A. 2022.
\newblock An Agnostic Approach to Federated Learning with Class Imbalance.
\newblock In \emph{Proceedings of the International Conference on Learning Representations (ICLR)}.

\bibitem[{Shoham et~al.(2019)Shoham, Avidor, Keren, Israel, Benditkis, Mor-Yosef, and Zeitak}]{shoham2019overcoming}
Shoham, N.; Avidor, T.; Keren, A.; Israel, N.; Benditkis, D.; Mor-Yosef, L.; and Zeitak, I. 2019.
\newblock Overcoming Forgetting in Federated Learning on Non-{IID} Data.
\newblock \emph{arXiv preprint arXiv:1910.07796}.

\bibitem[{Tan et~al.(2023)Tan, Liu, Long, Jiang, Lu, and Zhang}]{tan2023federated}
Tan, Y.; Liu, Y.; Long, G.; Jiang, J.; Lu, Q.; and Zhang, C. 2023.
\newblock Federated Learning on Non-{IID} Graphs via Structural Knowledge Sharing.
\newblock In \emph{Proceedings of the AAAI Conference on Artificial Intelligence}, 9953--9961.

\bibitem[{Tan et~al.(2022)Tan, Long, Liu, Zhou, Lu, Jiang, and Zhang}]{tan2022fedproto}
Tan, Y.; Long, G.; Liu, L.; Zhou, T.; Lu, Q.; Jiang, J.; and Zhang, C. 2022.
\newblock {FedProto}: Federated Prototype Learning Across Heterogeneous Clients.
\newblock In \emph{Proceedings of the AAAI Conference on Artificial Intelligence}, 8432--8440.

\bibitem[{Tang et~al.(2024)Tang, Han, Cai, Yu, Zhou, Oseni, and Das}]{tang2024personalized}
Tang, T.; Han, Z.; Cai, Z.; Yu, S.; Zhou, X.; Oseni, T.; and Das, S.~K. 2024.
\newblock Personalized Federated Graph Learning on {Non-IID} Electronic Health Records.
\newblock \emph{IEEE Transactions on Neural Networks and Learning Systems}, 35(9): 11843--11856.

\bibitem[{Tang and Liang(2024)}]{tang2024credit}
Tang, Y.; and Liang, Y. 2024.
\newblock Credit Card Fraud Detection Based on Federated Graph Learning.
\newblock \emph{Expert Systems with Applications}, 256: 124979.

\bibitem[{Veli{\v{c}}kovi{\'c} et~al.(2018)Veli{\v{c}}kovi{\'c}, Fedus, Hamilton, Li{\`o}, Bengio, and Hjelm}]{velivckovic2018deep}
Veli{\v{c}}kovi{\'c}, P.; Fedus, W.; Hamilton, W.~L.; Li{\`o}, P.; Bengio, Y.; and Hjelm, R.~D. 2018.
\newblock Deep Graph Infomax.
\newblock In \emph{Proceedings of the International Conference on Learning Representations (ICLR)}.

\bibitem[{Wang et~al.(2024)Wang, Zhang, Wang, Yuan, Cheng, Xu, and Yu}]{wang2024graphproxy}
Wang, J.; Zhang, L.; Wang, J.; Yuan, M.; Cheng, Y.; Xu, Q.; and Yu, B. 2024.
\newblock {GraphProxy}: Communication-Efficient Federated Graph Learning with Adaptive Proxy.
\newblock In \emph{Proceedings of the IEEE Conference on Computer Communications (INFOCOM)}, 2179--2188. IEEE.

\bibitem[{Weber et~al.(2019)Weber, Domeniconi, Chen, Weidele, Bellei, Robinson, and Leiserson}]{weber2019anti}
Weber, M.; Domeniconi, G.; Chen, J.; Weidele, D. K.~I.; Bellei, C.; Robinson, T.; and Leiserson, C.~E. 2019.
\newblock Anti-Money Laundering in {B}itcoin: Experimenting with Graph Convolutional Networks for Financial Forensics.
\newblock \emph{arXiv preprint arXiv:1908.02591}.

\bibitem[{Xiao and Wang(2023)}]{xiao2023triplets}
Xiao, C.; and Wang, S. 2023.
\newblock Triplets Oversampling for Class Imbalanced Federated Datasets.
\newblock In \emph{Joint European Conference on Machine Learning and Knowledge Discovery in Databases}, 368--383. Springer.

\bibitem[{Xie et~al.(2021)Xie, Ma, Xiong, and Yang}]{xie2021federated}
Xie, H.; Ma, J.; Xiong, L.; and Yang, C. 2021.
\newblock Federated Graph Classification over Non-{IID} Graphs.
\newblock \emph{Advances in Neural Information Processing Systems}, 34: 18839--18852.

\bibitem[{Yang et~al.(2024)Yang, Yu, Gao, Wang, Zhang, and Li}]{yang2024federated}
Yang, X.; Yu, H.; Gao, X.; Wang, H.; Zhang, J.; and Li, T. 2024.
\newblock Federated Continual Learning via Knowledge Fusion: A Survey.
\newblock \emph{IEEE Transactions on Knowledge and Data Engineering}, 36(8): 3832--3850.

\bibitem[{Zhang et~al.(2024)Zhang, Long, Zhou, Zhang, Yan, and Yang}]{zhang2024gpfedrec}
Zhang, C.; Long, G.; Zhou, T.; Zhang, Z.; Yan, P.; and Yang, B. 2024.
\newblock {GPFedRec}: Graph-Guided Personalization for Federated Recommendation.
\newblock In \emph{Proceedings of the 30th ACM SIGKDD Conference on Knowledge Discovery and Data Mining}, 4131--4142.

\bibitem[{Zhang et~al.(2023{\natexlab{a}})Zhang, Hua, Wang, Song, Xue, Ma, and Guan}]{zhang2023fedcp}
Zhang, J.; Hua, Y.; Wang, H.; Song, T.; Xue, Z.; Ma, R.; and Guan, H. 2023{\natexlab{a}}.
\newblock {FedCP}: Separating Feature Information for Personalized Federated Learning via Conditional Policy.
\newblock In \emph{Proceedings of the 29th ACM SIGKDD Conference on Knowledge Discovery and Data Mining}, 3249--3261.

\bibitem[{Zhang et~al.(2023{\natexlab{b}})Zhang, Li, Qi, and He}]{zhang2023survey}
Zhang, J.; Li, C.; Qi, J.; and He, J. 2023{\natexlab{b}}.
\newblock A Survey on Class Imbalance in Federated Learning.
\newblock \emph{arXiv preprint arXiv:2303.11673}.

\bibitem[{Zhang et~al.(2021)Zhang, Yang, Li, Sun, and Yiu}]{zhang2021subgraph}
Zhang, K.; Yang, C.; Li, X.; Sun, L.; and Yiu, S.~M. 2021.
\newblock Subgraph Federated Learning with Missing Neighbor Generation.
\newblock \emph{Advances in Neural Information Processing Systems}, 34: 6671--6682.

\end{thebibliography}

\section{Appendix}
\subsection{A. Derivation of $\mathcal{L}^{G\_{GAN}}_m$}
\label{app-derivation}
The derivation of the adversarial and diversity loss function for the hierarchical GAN, denoted as $\mathcal{L}^{G\_{GAN}}_m$, is founded on the principles established by the multi-agent diverse generative adversarial network (MAD-GAN) \cite{ghosh2018multi}. The core concept of MAD-GAN involves a multi-agent architecture with multiple generators and a single discriminator. To mitigate mode collapse, the MAD-GAN discriminator is trained not only to distinguish real samples from generated ones, but also to identify the source generator of any given fake sample. 

Our work adapts and extends the MAD-GAN framework through several key modifications. Following a hierarchical clustering process, each cluster $c_k$ is assigned a specific set of generators, $\{G_m|m\in c_k\}$,  and a corresponding discriminator, $D_k$. The architecture of the discriminator is augmented with an additional $H$-dimensional softmax layer. This layer is responsible for determining the probability that a given input corresponds to each class-specific feature prototype. This design compels each generator to produce samples that are not only realistic but also statistically distinct from those of its peers, thereby encouraging diverse exploration of the data manifold.

The training dynamics are governed by an adversarial objective. For the discriminator $D_k$, the objective is to align its predicted probability distribution for a generated sample with the client's true feature prototype distribution ($P(y|m)$). Conversely, the objective for a generator $G_m$ is to produce a distribution $P(y|G_m)$ that the discriminator cannot distinguish from the true data distribution $P(y|m)$, which is approximated by the client's local prototypes ($\mathcal{LP}_m$). Therefore, the objective for generator $G_m$ is to minimize the following loss function:
\begin{multline}
    \mathcal{L}_m^{G\_{GAN}} = \mathbb{E}_{x \sim P(y|m)} \log D_{k}(y|x) \\
    + \sum_{m'\in c_k} \mathbb{E}_{x \sim P(y|G_{m'})} \log(1 - D_{k}(y|x)),
\end{multline}
where $P(y|G_{m'})$ is the distribution produced by generator $G_{m'}$. The optimal discriminator $D^*_{k}$ is given by:
\begin{equation}
D^*_{k}(y|x) = \frac{P(y|m)}{P(y|m) + \sum_{m'\in c_k} P(y|G_{m'})}.
\end{equation}
 
By substituting the optimal discriminator $D^*_k$ into the generator's loss function, the objective can be expressed as:
{\small
\begin{multline}
    \mathcal{L}^{G\_{GAN}}_m = \mathbb{E}_{x \sim P(y|m)} \log \left[\frac{P(y|m)}{P(y|m)+\sum_{m'\in c_k} P(y|G_{m'})}\right] \\
    + \sum_{m'\in c_k} \mathbb{E}_{x \sim P(y|G_{m'})} \log \left[\frac{\sum_{m'\in c_k} P(y|G_{m'})}{P(y|m)+\sum_{m'\in c_k} P(y|G_{m'})}\right].\label{}
\end{multline}
}

This expression can be reformulated using the Kullback-Leibler (KL) divergence. Let the true data distribution be $P_{data}=P(y|m)$ and the aggregated model distribution be $P_{model}=\sum_{m'\in c_k}P(y|G_{m'})$. Defining the average mixture distribution as $P_{mix}=\frac{P_{data}+P_{model}}{2}$, the objective can be shown to be equivalent to twice the Jensen-Shannon divergence (JSD) between $P_{data}$ and $P_{model}$, up to a constant:
\begin{multline}
    \mathcal{L}^{G\_{GAN}}_m = \int P_{data} \log \frac{P_{data}}{P_{data}+ P_{model}} dx \\
    + \int P_{model} \log \frac{P_{model}}{P_{data} + P_{model}} dx \\
    = \int P_{data} \log \frac{P_{data}}{2 P_{mix}} dx + \int P_{model} \log \frac{P_{model}}{2 P_{mix}} dx \\
    = KL(P_{data} || P_{mix}) + KL(P_{model} || P_{mix}) - 2\log(2).
\end{multline}

The final form of the loss function used in this work is a specific instantiation of this objective, adapted for our multi-generator context. It incorporates two key components: First, a weighting factor, $H$ (the number of classes), is used to balance the contributions of the multiple generated distributions against the single true data distribution. This appropriately scales the penalty on the model's divergence. Second, a constant term related to entropy is included to normalize the objective and stabilize the training process by establishing a consistent baseline. This complete objective, expressed in terms of KL divergence, results in the following expression:
{\small
\begin{multline}
    \mathcal{L}^{G\_{GAN}}_m = \text{KL}\left(P(y|m) \| \frac{P(y|m)+\sum_{m'\in c_k} P(y|G_{m'})}{2}\right) \\
    + H \text{KL} \left(\sum_{m'\in c_k} P(y|G_{m'}) \| \frac{P(y|m)+\sum_{m'\in c_k} P(y|G_{m'})}{2}\right) \\
    - (H+1) \log (H+1) + H \log (H).
\end{multline}
}

This objective effectively incentivizes each generator's output distribution to converge towards a mixture of the true data distribution and the distributions of its peers, thereby fostering the desired ``cooperative competition."

\section{B. Datasets}\label{app-datasets}
Here we provide detailed descriptions of the four public graph datasets utilized in our evaluation. A primary characteristic shared by these datasets is significant class imbalance, making them ideal for assessing model performance in scenarios with underrepresented minority classes. The key statistics for these datasets are summarized in Table \ref{app-statistics}. The datasets, corresponding to the abbreviations used in the main text, are as follows:

\begin{itemize}
    \item \textbf{EllipticBitcoin (Elliptic)} \cite{weber2019anti}: This is a real-world graph of Bitcoin transactions designed for anomaly detection. Nodes represent transactions, edges denote the flow of currency, and the task is to classify transactions as ``licit" or ``illicit", with the illicit class constituting the minority.
    %\footnote{https://data.pyg.org/datasets/elliptic} 
    \item \textbf{Twitch (Twitch)} \cite{rozemberczki2021multi}: This dataset is a social network of streamers from the Twitch platform. Nodes represent individual streamers, and edges signify mutual follower relationships. The classification task is to identify streamers engaging in violative behavior, who constitute a minority of the user base.
    %\footnote{https://graphmining.ai/datasets/ptg/twitch}
    
    \item \textbf{FacebookPagePage (Facebook)} \cite{rozemberczki2021multi}: This dataset represents a network of official Facebook pages, where nodes are the pages and edges indicate mutual ``likes". The task is to classify pages with labels for page recommendation to users.
    %\footnote{https://graphmining.ai/datasets/ptg/facebook.npz}
    
    \item \textbf{Actor (Actor)} \cite{pei2020geom}: This is a co-actor network derived from film collaborations. Nodes represent actors, with features extracted from Wikipedia, and an edge connects two actors if they appeared in the same film. This task involves a five-way classification of actors, where one of the five groups is the designated minority class, leading to a notable class imbalance.
    %\footnote{https://raw.githubusercontent.com/graphdml-uiuc-jlu/geom-gcn/master}
    
\end{itemize}

For our experimental simulations, we partition these graphs to emulate a distributed setting. A number of clients are sampled for each dataset, with each client holding a subgraph containing 400 to 1200 nodes. The partitioning is performed in a manner that preserves the inherent class imbalance of the original dataset within each client's local data partition.

\begin{table}[t!]
\centering

\label{tab:dataset}

\begin{tabular}{ccccc}
\toprule
Statistics & Elliptic &  Twitch & Facebook & Actor  \\
\midrule
Nodes & 203,769 &   9,498 &22,470& 7,600  \\
Edges & 234,355 &   315,774 & 342,004 & 30,019  \\
Features & 165 & 128 & 128 & 932  \\
Classes & 2 & 2 & 4 & 5  \\
Client & 12 &  8 & 16 & 8  \\
Minority & 26.6\% & 38.9\% & 15.1\% & 11.0\%  \\
\bottomrule
\end{tabular}
\caption{The statistics of datasets.}
\label{app-statistics}
\end{table}

\begin{table*}[ht!]
    \centering
    \small
    % 增加了一列 (l)，并将 cmidrule 的列号向右移动了一位
    \begin{tabular}{ccccccccc}
    \toprule
    \multirow{2}{*}{Dataset} & \multirow{2}{*}{HC} & \multirow{2}{*}{GAN} & \multirow{2}{*}{MI loss} & \multirow{2}{*}{MIGMA} & \multicolumn{2}{c}{Accuracy} & \multicolumn{2}{c}{Recall} \\
    \cmidrule(lr){6-7} \cmidrule(lr){8-9}
    & & & & & Overall & Minority & Overall & Minority \\
    \midrule
    % --- 数据集 1 ---
    \multirow{4}{*}{Elliptic} & \checkmark &       &       &       & 0.871 & 0.695 & 0.816 & 0.696 \\
    & \checkmark & \checkmark &       &       & \textbf{0.895} & 0.730 & 0.842 & 0.729 \\
    & \checkmark & \checkmark & \checkmark &       & 0.891 & \textbf{0.744} & 0.846 & 0.729 \\
    & \checkmark & \checkmark & \checkmark & \checkmark & 0.894 & 0.740 & \textbf{0.849} & \textbf{0.733} \\
    \midrule
    % --- 数据集 2 ---
    \multirow{4}{*}{Twitch} & \checkmark &       &       &       & 0.573 & 0.473 & 0.543 & 0.456 \\
    & \checkmark & \checkmark &       &       & 0.561 & \textbf{0.587} & 0.560 & \textbf{0.551} \\
    & \checkmark & \checkmark & \checkmark &       & 0.588 & 0.512 & 0.570 & 0.508 \\
    & \checkmark & \checkmark & \checkmark & \checkmark & \textbf{0.597} & 0.515 & \textbf{0.582} &  0.520 \\
    \midrule
    % --- 数据集 3 ---
    \multirow{4}{*}{Actor} & \checkmark &       &       &       & 0.312 & 0.224 & 0.306 & 0.227 \\
    & \checkmark & \checkmark &       &       & 0.301 & 0.262 & 0.293 & 0.259 \\
    & \checkmark & \checkmark & \checkmark &       & 0.324 & 0.293& 0.294 & \textbf{0.314} \\
    & \checkmark & \checkmark & \checkmark & \checkmark & \textbf{0.327} & \textbf{0.297} & \textbf{0.319} & 0.297 \\
    \bottomrule
    \end{tabular}
    \caption{Ablation study on other datasets.}
    \label{app-ablation-study-performance}
\end{table*}

\section{C. Evaluation Metrics}\label{app-metrics}
Our evaluation relies on four distinct metrics to provide a comprehensive assessment of model performance.
\begin{itemize}
    \item \textbf{Overall accuracy}: This standard metric represents the proportion of all instances that are correctly classified across all classes. It provides a general measure of the model's correctness.
    \item \textbf{Minority accuracy}: This metric specifically assesses the model's performance on underrepresented classes. It is calculated as the average accuracy over all pre-defined minority classes, offering insight into how well the model handles infrequent data.
    \item \textbf{Overall recall}: This is the average recall score of all classes, measuring the overall ability of the model to avoid false negatives.
    \item \textbf{Minority recall}: This metric evaluates the model's sensitivity towards the minority classes. It is computed as the average recall across all minority classes, specifically gauging the model's effectiveness at identifying positive instances within these underrepresented groups.
\end{itemize}

\begin{figure*}[ht!]
\centering
\includegraphics[width=0.75\textwidth]{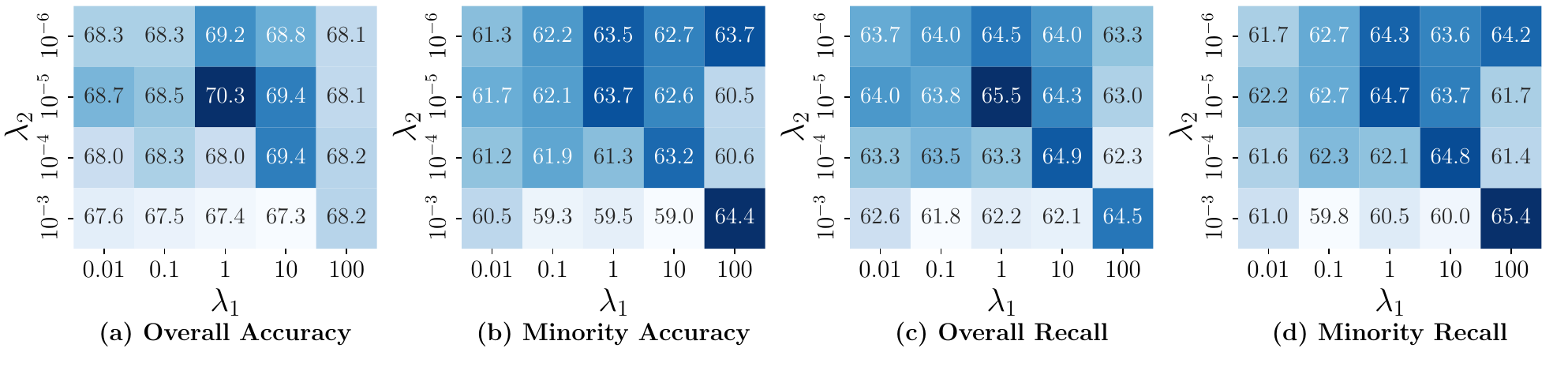} 
\caption{The results of hyperparameter $\lambda_1$ and $\lambda_2$ experiments of GraphFedMIG on Facebook.}
\label{app-hyperparameter-analysis-lambda}
\end{figure*}

\section{D. Baselines}\label{app-baselines}
Our experiments include a comparison against four baseline methods. The details of each are provided below.
\begin{itemize}
    \item \textbf{Local}: This baseline represents a non-federated approach where each client trains its GNN model exclusively on its own local data. There is no inter-client communication or model aggregation involved. This method serves as a performance lower bound, illustrating the results of isolated training.
    
    \item \textbf{FedAvg} \cite{mcmahan2017communication}: This is the canonical algorithm for federated learning. In this framework, clients independently train their local models, which are then periodically sent to a central server. The server aggregates these models weighted by the number of local samples to produce an updated global model that is subsequently distributed back to the clients.
    
    \item \textbf{FL+HC} \cite{briggs2020federated}: This method represents a clustered federated learning approach. It first performs hierarchical clustering on the clients, using the cosine similarity of their local data statistics as the grouping mechanism. Following the formation of clusters, the standard FedAvg protocol is then applied independently within each cluster of clients.
    
    \item \textbf{FedSpray} \cite{fu2024federated}: This baseline is a state-of-the-art federated graph learning method. To enhance privacy and performance, FedSpray introduces the concept of sharing structural proxies derived from local subgraphs instead of raw node features. These proxies are designed to encapsulate both structural and feature information while mitigating the impact of potentially adverse neighborhood information.
    
\end{itemize}

\section{E. Implementation Details}\label{app-implementation}

Our generator is a two-layer GraphSAGE model with a hidden dimension of 64. The discriminator consists of three linear layers followed by a final softmax activation, with the output dimension matching the number of classes in the dataset. For model optimization, we employ the Adam optimizer \cite{kinga2015method} with a learning rate of 0.01. The federated learning process is conducted for a total of 100 communication rounds. For all clustering-based baselines, cosine similarity is utilized as the similarity measure, denoted by $S(\cdot,\cdot)$. All methods are implemented using the PyTorch framework, and experiments are executed on a single NVIDIA RTX 3060 GPU. To ensure statistical reliability, all reported results represent the average of three independent runs performed with different random seeds. 
%The code is available at https://github.com/NovaFoxjet/GraphFedMIG.

\section{F. Ablation Study}

Table \ref{app-ablation-study-performance} presents the results of further ablation studies on the Elliptic, Twitch, and Actor datasets, which validate the efficacy of GraphFedMIG's components. On Elliptic and Actor, the results mirror our main findings: the progressive addition of the GAN and MI loss modules significantly boosts performance for the minority class, compensating for the weakness of the baseline HC model. The Twitch dataset presents a unique dynamic where the HC+GAN configuration yields the best minority results, suggesting the generative component alone can be highly effective for certain data distributions. Nevertheless, the complete GraphFedMIG model still secures the highest overall accuracy and recall on Twitch, highlighting its robustness in achieving a superior balance. Collectively, these results confirm the value of each component and the effectiveness of our integrated design in delivering strong, well-balanced performance across diverse datasets.

%\begin{figure*}[t!]
%\centering
%\includegraphics[width=0.9\textwidth]{Main/figures/app-convergence.pdf} 
%\caption{Model convergence comparison.}
%\label{app-convergence}
%\end{figure*}

\begin{figure}[t!]
\centering
\includegraphics[width=0.75\columnwidth]{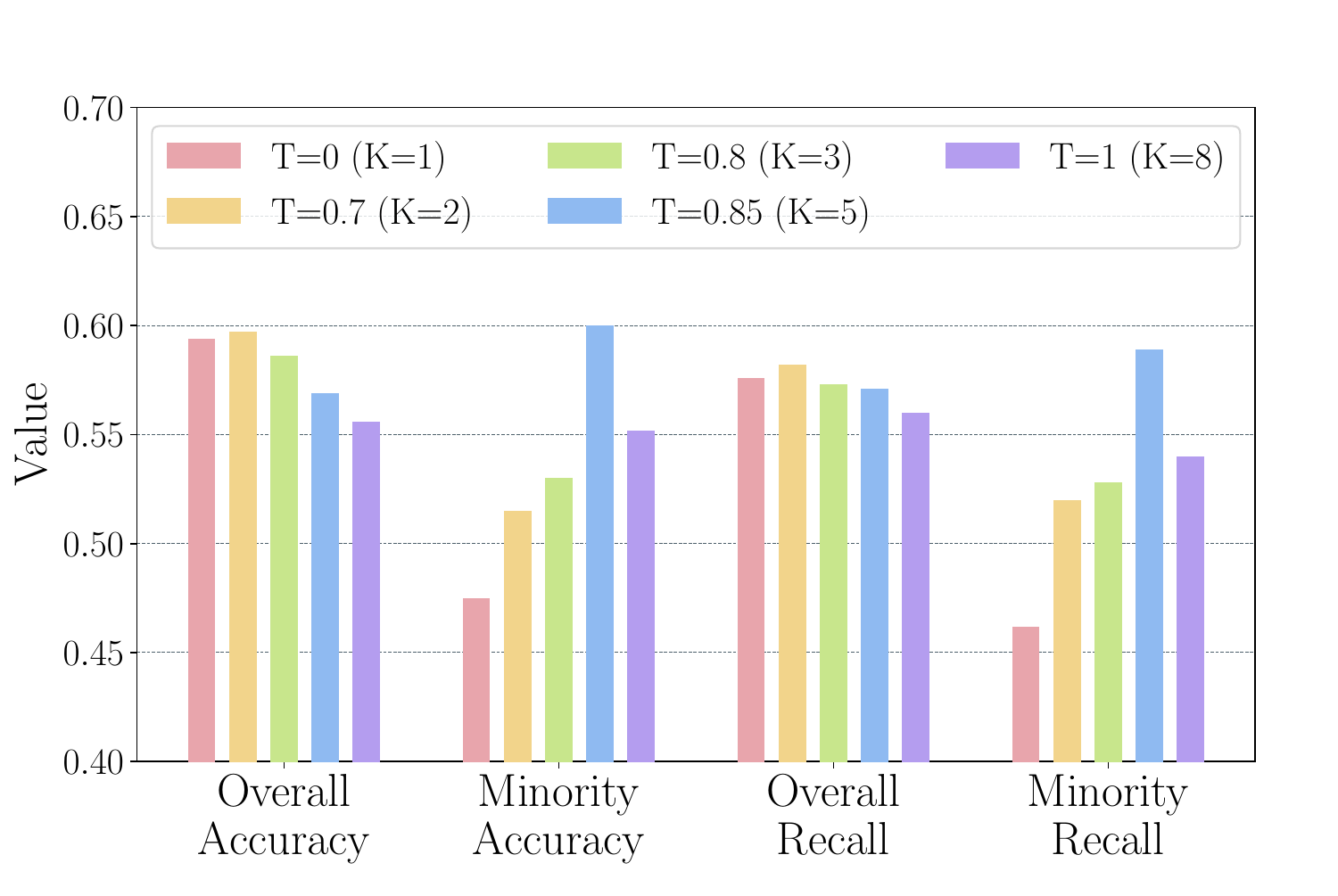} 
\caption{The results of the hyperparameter $T$ experiments of GraphFedMIG on Twitch.}\label{app-hyperparameter-analysis-T}
\end{figure}

\begin{figure}[t!]
\centering
\includegraphics[width=0.75\columnwidth]{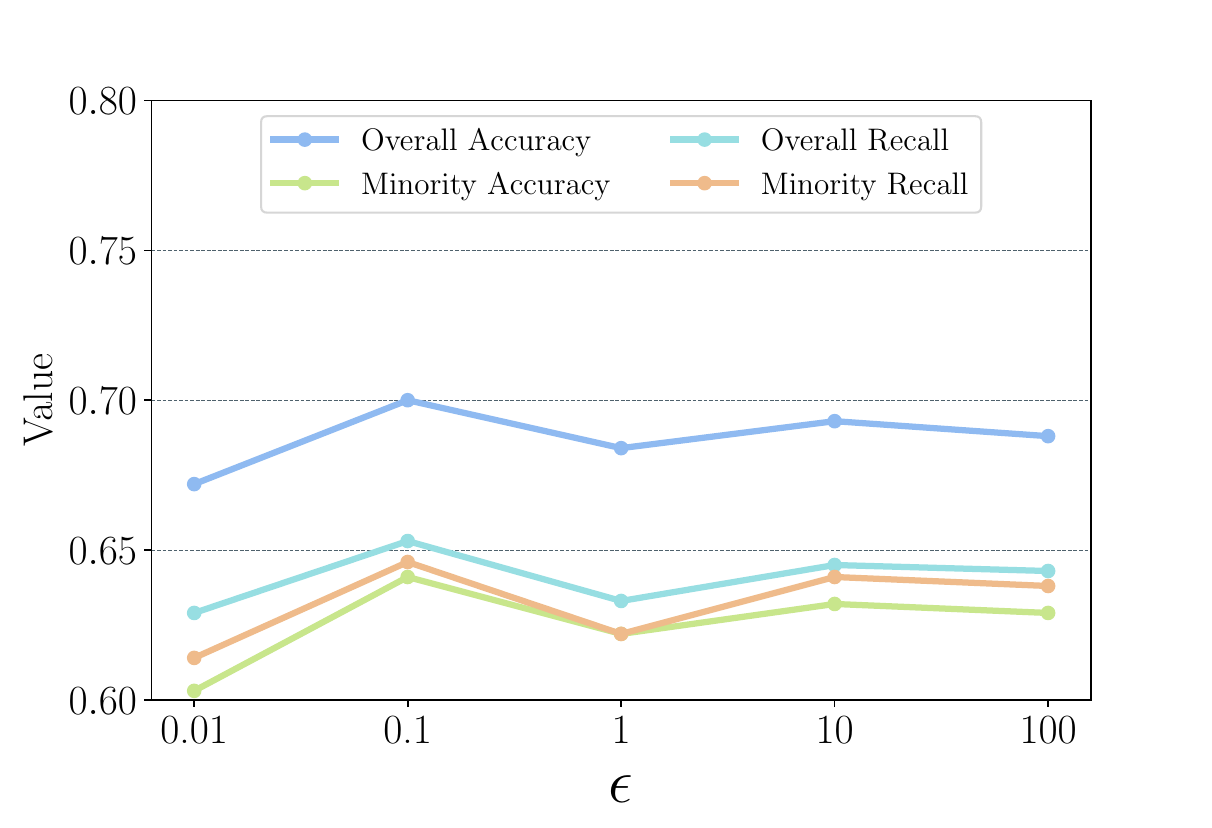} 
\caption{The results of the differential privacy experiments of GraphFedMIG on Facebook.}\label{app-privacy-protection}
\end{figure}

\section{G. Hyperparameter Analysis}

\subsubsection{Influence of $\lambda_1$ and $\lambda_2$.} 
Figure \ref{app-hyperparameter-analysis-lambda} presents the performance heatmap of GraphFedMIG on Facebook with different values of $\lambda_1$ and $\lambda_2$.
$\lambda_1$ and $\lambda_2$ control the generator's exploration for sample diversity and its fidelity to each client's local data distribution, respectively.
Figure \ref{app-hyperparameter-analysis-lambda} reveals that model performance is not dictated by the absolute values of $\lambda_1$ and $\lambda_2$, but by their relative ratio. Optimal results are consistently achieved when the ratio $\lambda_1/\lambda_2$ is approximately $10^5$. This optimal balance defines a diagonal band of high performance on the heatmaps. This suggests an ideal trade-off where the drive for sample diversity is significantly stronger than the constraint for fidelity.

Performance degrades when this optimal ratio is disrupted. If the ratio is substantially larger than $10^5$, the excessive focus on diversity slightly compromises overall performance. Conversely, if the ratio is substantially smaller, the overpowering fidelity constraint stifles exploration, causing a sharp drop in performance, particularly for minority-class metrics.

\subsubsection{Influence of $T$.}

We analyze the impact of the similarity threshold $T$, which controls the granularity of our client clustering by determining the final number of clusters $K$. A low threshold produces fewer and larger clusters, while a high threshold yields a more fine-grained partitioning of clients.
As shown in Figure \ref{app-hyperparameter-analysis-T}, $T$ reveals a fundamental trade-off between performance on minority classes and overall performance across both accuracy and recall. A low threshold results in poor minority performance, despite yielding high overall metrics. Conversely, a high threshold like $T=0.85$ $(K=5)$ maximizes minority performance, boosting minority accuracy and recall to their peaks. However, this comes at the cost of overall performance, where both accuracy and recall decrease.

\section{H. Differential Privacy Protection}

To ensure the privacy protection performance of GraphFedMIG, we incorporate $(\epsilon, \delta)$-differential privacy using the Gaussian mechanism \cite{abadi2016deep}. This approach is applied to the generated data that clients transfer during the federated learning process, safeguarding against potential information leakage. The Gaussian mechanism achieves privacy by adding noise drawn from a Gaussian distribution. The standard deviation $\delta$ of this noise is determined by the privacy budget $(\epsilon, \delta)$ and the function's L2 sensitivity $\Delta f$. The relationship is defined as:

\begin{equation}
\sigma_\text{noise} = \frac{\Delta f \sqrt{2 \ln\left(\frac{1.25}{\delta}\right)}}{\epsilon}.
\end{equation}

$\Delta f$ quantifies the maximum change in the function's output when applied to adjacent datasets $(\mathcal{D},\mathcal{D}^\prime)$ that differ by a single element:

\begin{equation}
\Delta f = \max_{\mathcal{D},\mathcal{D}^\prime} \left\| f(\mathcal{D}) - f(\mathcal{D}^\prime) \right\|,
\end{equation}
where the privacy budget $\epsilon$ controls the level of protection. A smaller $\epsilon$ provides stronger privacy guarantees. The parameter $\delta$ represents the probability that the privacy guarantee may not hold, and is conventionally set to a negligible value.

We conduct experiments on the Facebook dataset to analyze the impact of differential privacy on the performance of GraphFedMIG. The results, presented in Figure \ref{app-privacy-protection}, reveal a non-monotonic relationship between the privacy budget $\epsilon$ and model efficacy.
Notably, model performance does not simply degrade as the privacy budget becomes stricter (i.e., as $\epsilon$ decreases). Instead, the model achieves peak performance across all metrics at an intermediate privacy level of $\epsilon=0.1$. This counter-intuitive result suggests that a moderate level of noise introduced by the privacy mechanism acts as a form of regularization. This likely mitigates overfitting in the generator module, leading to better generalization compared to models with very little noise (e.g., at $\epsilon=100$).

\begin{table}[t!]
\centering

\begin{tabular}{cccc}
\toprule
FedAvg & FL+HC & FedSpray & GraphFedMIG \\
\midrule
0.049MB& 0.054MB & 0.082MB & 0.089MB \\
\bottomrule
\end{tabular}
\caption{Comparison of communication overhead.}\label{app-communication}
\end{table}

\section{I. Communication Overhead}\label{app-efficiency}
As detailed in Table \ref{app-communication}, we analyze the communication overhead of our proposed framework. Although the cost of GraphFedMIG (0.089 MB per round) is quantitatively higher than the baselines, it is crucial to note that all methods operate within the same order of magnitude.

This modest increase in overhead is inherent to our generative approach, which requires the exchange of additional components (the generator, discriminator, and feature prototypes) to effectively model heterogeneous client data. We strategically design these components to be lightweight. For instance, the feature prototypes are low-dimensional, and the discriminator is significantly less complex than the base GNN. Consequently, the primary additional cost is attributed to the generator parameters, which are essential for learning the underlying data distributions. Hence, we posit that this marginal communication overhead is a deliberate and justified trade-off for the substantial performance improvements our method achieves.

\end{document}